\documentclass[a4paper]{article}       
\pdfoutput=1
\usepackage[utf8x]{inputenc}
\usepackage{ucs}
\usepackage{amsmath}
\usepackage{amsfonts}
\usepackage{amssymb}
\usepackage{setspace}
\usepackage{graphicx}
\usepackage{placeins}
\usepackage{url}
\usepackage{caption}
\usepackage{subcaption}
\usepackage{tikz}
\usepackage{pgfplots}
\usepackage{adjustbox}
\newtheorem{mydef}{Definition}

\begin{document}

\title{From visual words to a visual grammar: using language modelling for image classification}


\author{Antonio Foncubierta-Rodr\'iguez         \and
        Henning M\"uller \and
        Adrien Depeursinge
}

\maketitle
\begin{abstract}
The Bag--of--Visual--Words (BoVW)  is a visual description technique that aims at shortening the semantic gap by partitioning a low--level feature space into regions of the feature space that potentially correspond to visual concepts and by giving more value to this space. 
In this paper we present a conceptual analysis of three major properties of language grammar and how they can be adapted to the computer vision and image understanding domain based on the bag of visual words paradigm. 
Evaluation of the visual grammar shows that a positive impact on classification accuracy and/or descriptor size is obtained when the technique are applied when the proposed techniques are applied. 
\end{abstract}
\section{Introduction}
Image retrieval and classification has been an extremely active research domain with hundreds of publications in the past 20 years~\cite{MMB2004,ARN2011,THI1999}.
Initiatives like the PASCAL Visual Object Class (VOC) challenge~\cite{EEV2015} and ImageCLEF~\cite{MCD2010}  have attracted many research groups to compare their methods for retrieval and classification tasks. 

The Bag--of--Visual--Words (BoVW)  is a visual description technique that aims at shortening the semantic gap by partitioning a low--level feature space into regions of the features space that potentially correspond to visual concepts. 
These regions are called visual words in an analogy to text--based retrieval and the bag of words approach. 
An image can be described by assigning a visual word to each of the feature vectors that describe local regions of the images (either via a dense grid sampling or interest points), and then representing the set of feature vectors by a histogram of the visual words. 
One of the most interesting characteristics of the BoVWs is that the set of visual words is created based on the actual data and therefore only concepts present in the data will be part of the visual vocabulary~\cite{GrL2011}.
The creation of the vocabulary is normally based on a clustering method (e.g. k--means, DENCLUE) to identify local clusters in the feature space and then assigning a visual word to each of the cluster centers. 
This has been investigated previously, either by searching for the optimal number of visual words~\cite{FDM2012}, by using various clustering algorithms~\cite{HiG2007} instead of k--means or by selecting interest points to obtain the features~\cite{HDB2011}. 
Although the BoVW is widely used in the literature~\cite{AGK2011,MGE2012} there is a strong performance variation within similar experiments when considering different vocabulary sizes~\cite{FDM2012}.

Fisher Vectors~\cite{KVJ2011,SPM2013} have been proposed to overcome some of the limitations of the BoVW, improving the classification accuracy or retrieval precision. In~\cite{CLV2011} Chatfield et al. performed exhaustive comparisons between various feature encoding methods and histogram--based BoVW. 
In the reported results, Fisher Vectors perform better in terms of precision and accuracy than the baseline BoVW. However, the improvements in performance were obtained at the cost of descriptors that are 64\% to 412\% longer than those used by the baseline. 

On the other hand, some language modelling concepts have already been exported from text to BoVW--based techniques, such as stop--words~\cite{TCG2008,TZZ2011,FGM2013,FGM2014}. These methods have proven that classification accuracy can be improved by removing \textit{noisy} words rather than by increasing the dimensionality of the descriptor.
However, the use of language modelling techniques is limited to a small set of situations and even though results are promising no generic model is given to take advantage of this application.

In this article a novel, generic, language modelling--based model is proposed. The Visual Grammar model provides tools that allow improving image understanding by computer--based techniques without increasing the size of the descriptor using linguistic concepts that are easily understood by humans: topics, meaningfulness, synonymy and polysemy. 

The rest of the article is organized as follows: section~\ref{sec:Notation} sets the notation used in the rest of the article, section~\ref{sec:VisualGrammar} introduces the Visual Grammar Model and its three transformations (meaningfulness in section~\ref{sec:Meaningfulness}, synonymy in section~\ref{sec:Synonymy} and polysemy in section~\ref{sec:Polysemy}), experimental evaluation using the PASCAL VOC 2007 and the ImageCLEF 2013 datasets is covered in section~\ref{sec:Evaluation} and discussion of results and conclusions are left for sections~\ref{sec:Discussion} and~\ref{sec:Conclusions}.
\section{Notation}
\label{sec:Notation}
In this paper we use terms from a variety of research domains, including image analysis, text retrieval, linguistics and machine learning. In this section we present the notation that we use, and define terms from a conceptual point of view. 
\begin{itemize}
\item A \emph{visual instance}, is the basic unit of visual information 
that we are interested in describing. It may correspond to a {2D} image, a video, a volumetric image or a region of interest in any of them. Formally, we will refer to visual instances using the letter $I$, indexed by $\left\lbrace 1,\dots,N_I \right\rbrace$. 
\item A \emph{collection} or \emph{corpus} is the set containing all visual instances: $\mathcal{I} = \left\lbrace I_1,\dots,I_{N_I} \right\rbrace$. 
\item A \emph{feature} is a measurable value of the visual properties of the visual instance, e.g. a filter response. 
Formally we refer to features using the letter $f$. Features can be grouped into \emph{feature vectors} $\mathbf{f}$ in a \emph{feature space} $S_f$
\item A \emph{visual word} $w$ is a specific region of the feature space $S_f$, created by the clustering of the feature space.
A visual word is defined to be an item from a vocabulary  $\mathcal{V} = \left\lbrace w_1,\dots,w_{N_W} \right\rbrace$.
\item The \emph{bag of visual words} of a visual instance $I_i$ is a multiset of $L_i$ elements where each item belongs to the vocabulary  $\mathcal{V}$. 
Each $I_i$ can be represented by the histogram of visual words, a $N_W$--dimensional vector where the $j$--th component is the multiplicity of the word $w_n$ in the visual instance $I_i$:  $ \mathbf{h_{i}} = (n_i(w_1),n_i(w_2),\dots,n_i(w_{N_W}))$, with $\sum_{n=1}^{N_W} n_i(w_n) = L_i $.  
\end{itemize}
\section{The visual grammar model}
\label{sec:VisualGrammar}
Representing visual information using a histogram of visual words poses the obvious question of how visual words are chosen in order to convey a meaningful description of the visual instance. 
It also requires optimizing the relative weights of words according to their importance, meaningfulness, ambiguity, etc. 
Weighting of word importance in text retrieval is a challenging area where various models have been proposed~\cite{YaP1997,Gre1998} with \emph{tf--idf}\footnote{TF--IDF refers to a word weighting scheme where the term frequency (tf) and the inverse document frequency (idf) are considered.} and BM25 being among the most popular ones~\cite{SaM1983,RVP1980,RoW1994}.

Visual words are often generated using a clustering method in a feature space populated with training data. Experimental results show that there is no optimal number of visual words for all image description tasks~\cite{FDM2012,MGE2012}. 
Larger vocabularies can produce smaller, more compact clusters that are able to model subtle differences among neighboring visual words. But they can also split meaningful clusters into various words with a similar meaning (synonyms), with a smaller weight in the histogram. 
On the other hand, smaller vocabularies merge words into a single large cluster that contains a mixture of all meanings (polysemy).

\label{sec:synonymyPolysemyIntroduction}
The cornerstone of this paper is to identify relations among visual words to improve image understanding. 
Identifying the topics present in a collection and quantifying word relevance for each of the topics is a first approximation to understanding word--level relations. Later, these relations are further analyzed in terms of the synonymy and polysemy concepts. 
\subsection{Visual topics}
In spoken or written language, not all words contain the same amount of information. 
Specifically, the grammatical class of a word is tightly linked to the amount of meaning it conveys. 
E.g. nouns and adjectives (open grammatical classes) can be considered more informative than prepositions and pronouns (closed grammatical classes).

Similarly, in a vocabulary of $N_W$ visual words generated by clustering a feature space populated with training data, not all words are useful to describe the appearance of the visual instances. 

From an information theoretical point of view, a bag of (visual) words containing $L_i$ elements can be seen as $L_i$ observations of a random variable $W$. The unpredictability or information content of the observation corresponding to the visual word $w_n$ is 
\begin{equation}
I(w_n)=log\left(\frac{1}{P(W=w_n)}\right) 
\label{eq:informationDef}
\end{equation}
This explains why nouns or adjectives contain, in general, more information than prepositions or pronouns. Those words belonging to a closed class are more probable than those belonging to a much richer class. According to Equation~\ref{eq:informationDef}, information is related to unlikelihood of a word. 

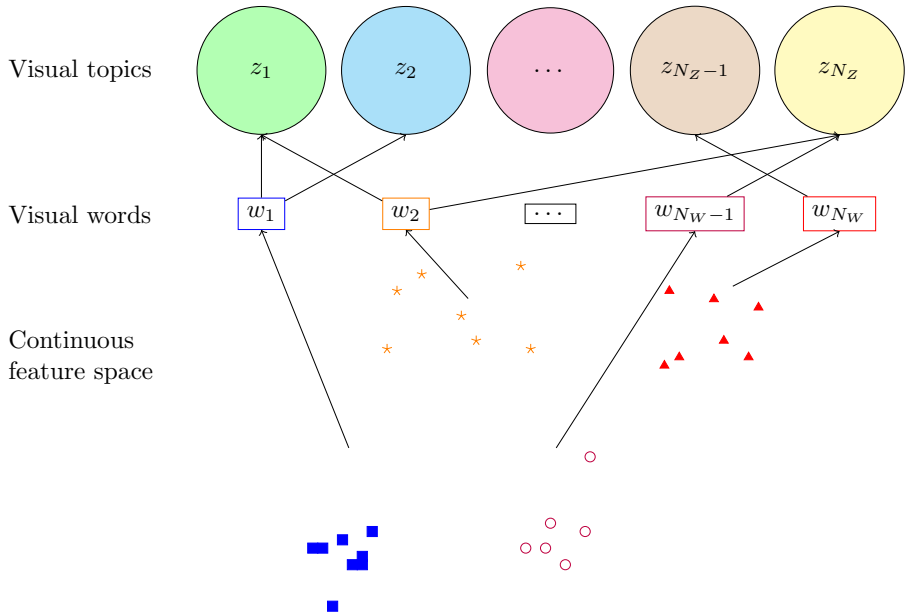
\begin{figure}
\centering
\adjustbox{max width=\linewidth}{
\begin{tikzpicture}[node distance=2cm,auto]
\begin{axis}[axis lines = none, xlabel = $f_1$, ylabel = {$f_2$}, xmin=0, xmax=1, scatter/classes={ 
  a={mark=square*,blue}, 
  b={mark=triangle*,red}, 
  c={mark=o,draw=purple},
  d={mark=star,draw=orange}}]
\addplot[scatter,only marks,scatter src=explicit symbolic] 
table[meta=label] 
{
x     y      label
0.1   0.125   a 
0.02  0.135   a 
0.06  0.1    a 
0.12  0.125   a 
0.12  0.13   a 
0.04  0.135   a 
0.08  0.14   a 
0.14  0.145   a %
0.71  0.225   b 
0.92  0.28   b 
0.83  0.285   b 
0.74  0.29   b 
0.73  0.245   b 
0.9   0.25    b 
0.85  0.26   b 
0.76  0.25    b %
0.55  0.16   c
0.53  0.125   c 
0.5   0.15    c 
0.45  0.135   c %
0.56  0.155   c
0.57  0.145   c 
0.58  0.19   c 
0.49  0.135   c %
0.21  0.33   d
0.32  0.275   d
0.44  0.305   d
0.19  0.29   d
0.24  0.300   d
0.35  0.26   d
0.46  0.255   d
0.17  0.255   d
};
\node at (axis cs:0.10,0.19) (Cluster1) {};
\node at (axis cs:0.35,0.28) (Cluster2) {};
\node at (axis cs:0.5,0.19) (Cluster3) {};
\node at (axis cs:0.85,0.29) (Cluster4) {};

\node at (axis cs:0.5,0.25) (arriba) {};

\end{axis}

\node [above of=arriba,draw=black] (mitad) {\ldots};
\node [left of=mitad, draw=orange] (VW2) {$w_2$};
\node [left of=VW2, draw=blue] (VW1) {$w_1$};
\node [right of=mitad, draw=purple] (VW3) {$w_{N_W-1}$};
\node [right of=VW3, draw=red] (VWN) {$w_{N_W}$};

\path[->] (Cluster1) edge node {} (VW1.south);
\path[->] (Cluster2) edge node {} (VW2.south);
\path[->] (Cluster3) edge node {} (VW3.south);
\path[->] (Cluster4) edge node {} (VWN.south);

\node[above of=VW1] (wsep) {};
\node[above of= VW1, circle, text centered,text width=1.5cm, draw=black, fill=green!30] (T1) {$z_1$};
\node[right of= T1, circle,  text centered,text width=1.5cm, draw=black, fill=cyan!30] (T2) {$z_2$};
\node[right of= T2, circle,  text centered,text width=1.5cm, draw=black, fill=magenta!30](T3) { \ldots};
\node[right of= T3, circle,  text centered,text width=1.5cm, draw=black, fill=brown!30] (T4) {$z_{N_Z-1}$};
\node[right of= T4, circle,  text centered,text width=1.5cm, draw=black, fill=yellow!30] (T5) {$z_{N_Z}$};

\path[->] (VW1) edge node {} (T1.south);
\path[->] (VW1) edge node {} (T2.south);
\path[->] (VW2) edge node {} (T5.south);
\path[->] (VW3) edge node {} (T5.south);
\path[->] (VW2) edge node {} (T1.south);
\path[->] (VWN) edge node {} (T4.south);

\node[left of=T1, text width=3cm ] (topiclayer) {Visual topics};
\node[below of=topiclayer, text width=3cm ] (wordlayer) {Visual words};
\node[below of=wordlayer, text width=3cm ] (continuous) {Continuous\\ feature space };

\end{tikzpicture}
  }
  \caption{Conceptual model of visual topics, words and features. Whereas continuous features are the most informative descriptors from an information theoretical point of view, visual words generalize feature points that are close in the feature space. We propose visual topics as a higher generalization level, modelling partially shared meanings among words.}
  \label{fig:visualTopicsDiagram}
\end{figure}

In a bag of visual words scheme for visual understanding it is important to use very specific words with high discriminative power. 
On the other hand, using very specific words alone does not always allow to establish and recognize similarities. 
This can be done by establishing a concept that generalizes very specific words that share similar meanings into a less specific \emph{visual topic}, as shown in Figure~\ref{fig:visualTopicsDiagram}.

In the definition of Probabilistic Latent Semantic Analysis (PLSA\footnote{PLSA is an extension of Latent Semantic Analysis (LSA)~\cite{Dee1990}, a language modelling technique that maps documents to a vector space of reduced dimensionality, called \emph{latent semantic space}, based on a Singular Value Decomposition (SVD) of the terms--documents occurrence matrix.} )~\cite{Hof2001}, Hofmann defines a generative model that states that the observed probability of a word or term occurring in a given document is linked to a latent or unobserved set of topics (also called aspects) in the text. 

Since it does not set any requirements on the nature of the low level features that yield these co--occurrence matrices (other than being discrete), the extension to visual words is straightforward.
PLSA in combination with visual words for classification and retrieval purposes was also applied in~\cite{BZM2006,EMU2012}. In~\cite{TCG2008} PLSA is proposed to remove noisy visual words. This approach is further extended with the concept of \emph{meaningfulness} in~\cite{FGM2013}, obtaining reductions of up to 92\% of the vocabulary size without significant effect on image classification accuracy. 
\begin{mydef}[PLSA--based visual topic]
A visual topic is an unobserved or latent variable $z \in \mathcal{Z} = \left\lbrace z_1,\dots,z_{N_Z} \right\rbrace$  so that the probability of observing the word $w_n$ in the visual instance $I_i$:
$$
P(w_n,I_i)=\sum_{j=1}^{N_Z}  P(w_n|z_j)P(z_j|I_i).
$$
\end{mydef}
The model is fit via the EM (Expectation--Maximization) algorithm. 
For the expectation step: 
\begin{equation}
P(z_j|I_i,w_n)=\frac{P(w_n|z_j)P(z_j|I_i)}{\sum_{j=1}^{N_Z} P(w_n|z_j)P(z_j|I_i)}.
\end{equation}
and for the maximization step:
\begin{eqnarray}
P(w_n|z_j)=\dfrac{\sum_{i=1}^{N_I} n(I_i,w_n)P(z_j|I_i,w_n)}{\sum_{m=1}^{N_W} \sum_{i=1}^{N_I} n(I_i,w_m)P(z_j|I_i,w_m)},
\\
P(z_j,I_i)=\dfrac{\sum_{m=1}^{M_W} n(I_i,w_m)P(z_j|I_i,w_m)}{n(I_i)}.
\end{eqnarray}
where $n(I_i,w_n)$ denotes the number of times the term $w_n$ occurred in the visual instance $I_i$; and $n(I_i)=\sum_k(I_i,w_n)$ refers to the total number of visual words in the visual instance $I_i$.

These steps are repeated until convergence or until a termination condition is met.
As a result, two probability matrices are obtained: the word--concept probability matrix $W_{N_W \times N_Z} = (P(w_n|z_j))_{n,j}$ and the concept--visual instance probability matrix $D_{N_Z \times N_W} = (P(z_j|I_i))_{j,i}$.
\subsection{Meaningfulness transformation}
\label{sec:Meaningfulness}
Arguably, the most obvious transformation is to weight words according to their meaningfulness. 
As a first approximation to topic--based word weighting, visual significance for each visual word/topic pair can be quantified. Following the ideas from~\cite{TCG2008,FGM2013} we define the visual significance of a word for a given topic. This quantifies how much a word belongs to a given topic. 
\begin{mydef}[Topic--based significance]
\label{sec:topicBasedSignificance}
Given a visual topic $z_j \in \mathcal{Z}$ and the set of probabilities $\mathcal{P} = \left\lbrace P(w_m|z_j) \right\rbrace \; \forall m = 1,\dots,N_W$, the significance of a word $w_n$ for the visual topic $z_j$ is defined as the ratio of elements in $\mathcal{P}$ with a lower conditional probability than $P(w_n|z_j)$:
$$
t_{n,j} = \dfrac{\vert \left\lbrace p \in \mathcal{P} \mid p \leq P(w_n|z_j) \right\rbrace \vert}{N_W}
$$
\end{mydef}
\begin{mydef}[Visual meaningfulness]
The visual meaningfulness of a visual word $w_n$ is its maximum topic--based significance level:  
$$
m_n = \left\lbrace  
\begin{array}{l}
\max_{j} \left\lbrace t_{n,j}\right\rbrace \ \mathrm{if}\ \max_{j} \left\lbrace t_{n,j}\right\rbrace \geq T_{meaning} \\
0 \ \mathrm{otherwise} \\
\end{array}
\right .
$$ 
\end{mydef} 

Given a meaningfulness threshold $T_{meaning}$, words that are not meaningful for any concept at this level can be removed from the visual word space, producing a \emph{meaningfulness--truncated feature space}.
This approach was tested in~\cite{FGM2013}, achieving reduction ratios of up to $92\%$ of the feature space with a limited cost in classification accuracy and retrieval precision. 

Instead of using a hard decision based on a meaningfulness threshold, a transformation can be defined to weight visual words according to their meaningfulness.
\begin{mydef}[Meaningfulness--transformed visual word space]
Let $\mathbf{h}$ be a histogram vector where each component represents the multiplicity of a visual word, and $\mathbf{M}$ a meaningfulness transformation matrix:
\begin{eqnarray}
\mathbf{h} & = & (n(w_1),n(w_2),\dots, n(w_{N_W}))^T
\\
\mathbf{M} & = & \left( 
\begin{array}{cccc}
m_1 & 0 & \cdots & 0 \\ 
0 & m_2 & \cdots & 0 \\ 
\vdots & \vdots & \ddots & \vdots \\ 
0 & 0 & \cdots & m_{N_W}
\end{array} 
 \right)
\end{eqnarray}

Then, the vector $\mathbf{h^M} =(n(w^M_1),n(w^M_2),\dots, n(w^M_{N_W}))^T $ is the histogram vector of visual words in the meaningfulness--transformed space.
\begin{eqnarray}
\mathbf{h^M} & = & \mathbf{M}\mathbf{h} \\
n(w^M_i) & = & m_i\cdot n(w_i)
\end{eqnarray}
\end{mydef}
\subsection{Synonymy transformation}
\label{sec:Synonymy}
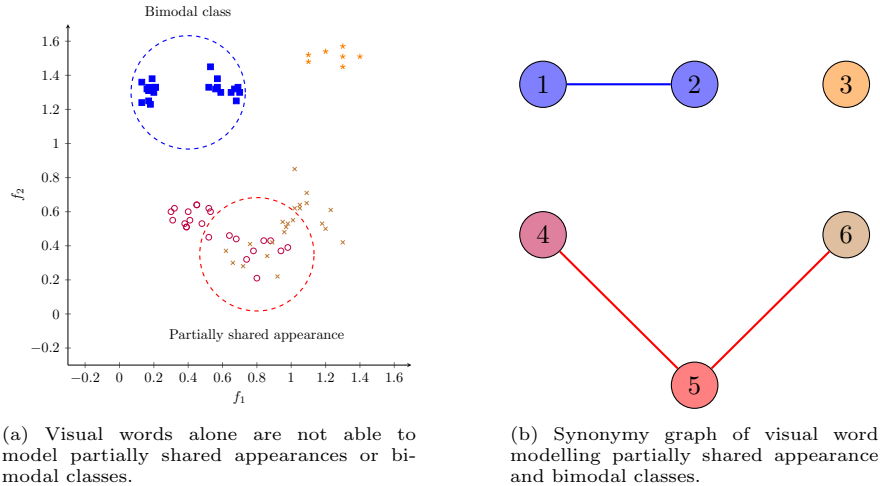
\begin{figure}[hbt]
\centering
\begin{subfigure}[t]{0.45\linewidth}
\centering
\adjustbox{max width=\linewidth}{
  \begin{tikzpicture}[node distance=2cm,auto]
\begin{axis}[width=10cm, height=10cm, axis lines = left, xlabel = $f_1$, ylabel = {$f_2$}, xmin=-0.3, xmax=1.7, ymin=-0.3, ymax=1.7, scatter/classes={ 
  a={mark=square*,blue}, 
  b={mark=triangle*,red}, 
  c={mark=o,draw=purple},
  d={mark=x,draw=brown},
  e={mark=star,draw=orange}}]
\addplot[scatter,only marks,scatter src=explicit symbolic] 
table[meta=label] 
{
x     y      label
0.20 1.30 a 
0.19 1.32 a
0.21 1.33 a
0.17 1.25 a
0.13 1.36 a
0.16 1.32 a
0.17 1.33 a
0.18 1.23 a
0.19 1.38 a
0.17 1.31 a
0.21 1.33 a
0.13 1.24 a
0.70 1.30 a 
0.49 1.42 a
0.69 1.33 a
0.57 1.38 a
0.65 1.30 a
0.56 1.32 a
0.57 1.33 a
0.68 1.25 a
0.59 1.30 a
0.67 1.32 a
0.52 1.33 a
0.53 1.45 a

0.80 0.21 c 
0.82 0.35 d 
0.84 0.43 c 
0.86 0.34 d
0.88 0.43 c
0.92 0.22 d 
0.94 0.37 c 
0.96 0.48 d
0.98 0.39 c
0.62 0.37 d 
0.64 0.46 c 
0.66 0.30 d
0.68 0.44 c
0.72 0.28 d 
0.74 0.32 c 
0.76 0.41 d
0.78 0.37 c

0.30 0.60 c 
0.31 0.55 c 
0.45 0.64 c 
0.38 0.53 c 
0.52 0.62 c 
0.39 0.51 c 
0.53 0.60 c 
0.40 0.60 c 
0.41 0.55 c 
0.45 0.64 c 
0.48 0.53 c 
0.32 0.62 c 
0.39 0.51 c 
0.52 0.45 c 

1.20 0.50 d
1.30 0.42 d 
1.01 0.55 d 
1.05 0.64 d 
1.18 0.53 d 
1.02 0.62 d 
1.09 0.71 d 
1.23 0.61 d 
1.05 0.62 d 
1.09 0.65 d 
0.95 0.54 d 
0.98 0.53 d 
0.89 0.42 d 
0.97 0.51 d 
1.02 0.85 d

1.30 1.51 e
1.10 1.52 e
1.20 1.54 e
1.30 1.57 e
1.40 1.51 e
1.10 1.48 e
1.30 1.45 e

};
\node at (axis cs: 0.2,1.3) (ClusterA1) {};
\node at (axis cs: 0.5,1.3) (ClusterA2) {};
\node at (axis cs: 0.4,1.3) (ClusterAcenter) {};
\node at (axis cs: 0.8,0.35) (ClusterB) {};
\node at (axis cs: 0.4,0.6) (ClusterC) {};
\node at (axis cs: 1.08,0.7) (ClusterD) {};
\node at (axis cs: 1.25,1.45) (ClusterE) {};

\end{axis}

\draw[thick, dashed, blue] (ClusterAcenter) circle (1.4cm);
\draw[thick, dashed, red] (ClusterB) circle (1.4cm);

\node [above of=ClusterAcenter] {Bimodal class};
\node [below of=ClusterB] {Partially shared appearance};

\end{tikzpicture}
  }
\caption{Visual words alone are not able to model partially shared appearances or bimodal classes.}
\label{fig:clustersAndClasses}
\end{subfigure}
\hspace{1cm}
\begin{subfigure}[t]{0.4\linewidth}
\centering
\adjustbox{max width=\linewidth}{
  \begin{tikzpicture}[node distance=2cm,auto]

\node[circle, draw=black, fill=blue!50] (clusterA1) {1};
\node[right of=clusterA1, circle, draw=black, fill=blue!50] (clusterA2) {2};
\node[right of=clusterA2, circle, draw=black, fill=orange!50] (clusterE) {3};
\node[below of=clusterA2] (centro) {};
\node[below of=clusterA1, circle, draw=black, fill=purple!50] (clusterC) {4};
\node[below of=centro, circle, draw=black, fill=red!50] (clusterB) {5};
\node[below of=clusterE, circle, draw=black, fill=brown!50] (clusterD) {6};

\path[thick,blue] (clusterA1) edge node {} (clusterA2);
\path[thick,red] (clusterC) edge node {} (clusterB);
\path[thick,red] (clusterD) edge node {} (clusterB);

\end{tikzpicture}
  }
\caption{Synonymy graph of visual word modelling partially shared appearance and bimodal classes. }
\label{fig:clustersAndSynonyms}
\end{subfigure}
\caption{A multi--class situation where classes can have various visual appearances (green squares distributed into two clusters) or partially share a visual pattern (purple and red stars and circles belonging to the same cluster). A graph can represent the synonymy relations among words when meaning is shared. }
\label{fig:clusters}
\end{figure}

As stated above, one of the key aspects of the bag of visual words approach is that the visual words are learnt from a training data set. 
If the visual word creation was controlled, words would be produced at the desired level of specificity: one word for each visual pattern to be distinguished. 
However, supervising the creation of visual words with class--based ground truth goes against the notion of learning the visual patterns present in the data independently from the classes. Furthermore, the number of visual patterns might be unknown and/or independent of the number of classes. E.g.: in a multi--class situation, two or more classes might partially share a visual pattern, or a single class might have several visual appearances. Figure~\ref{fig:clustersAndClasses} illustrates such a situation.

Synonymy is the property of two words that have the same meaning. 
Although it can be discussed that absolute synonymy might not exist, as the choice of one word over its equivalent already conveys meaning. 
This concept is known as paradigmatic relation, i.e. a word belongs to a \emph{paradigm} or group of words with similar meaning, and the choice of one over the other words from the same group is as informative as the shared meaning of the group. 
In Figure~\ref{fig:clustersAndSynonyms}, these relations are expressed using a graph: words that partially belong to the same paradigm are linked. 
Since words are not merged, information is preserved, and synonymy relations provide additional information.  

From a text analysis point of view, synonymy relations can be inferred from the distribution and association of words, topics and documents. 

The distributional hypothesis~\cite{Har1954,Dee1990,TuP2010} states that words with similar meanings occur in similar contexts in the corpus and therefore have a similar contextual distribution. 
In the bag of visual words approach, a context might be a complete visual instance (image, video, etc.) or a subregion of the visual instance. 
The use of the context as a subregion of a visual instance mimics the use of n--grams, where word occurrences are studied in contiguous groups. 
Choosing the size of the context is equivalent to choosing the length of the n--grams. 

The distributional hypothesis is one of the most extended for discovering semantic relations among words. However, synonymy is one step beyond the semantic relation, and introduces the notion of equivalence or complementarity. 
Therefore, if two visual words are synonyms or equivalent, it is very unlikely that they will be used together in the same context. 
Instead, they will probably have a complementary distribution. 

In~\cite{ChH1990} an information theoretic measure is defined for analyzing word associations in a document corpus. 
In a bag of visual words approach, we can use a similar definition to measure the \emph{associatedness} of two words. 
\begin{mydef}[Point--wise mutual information]
The point--wise mutual information or \emph{association ratio} of a pair of visual words $w_n, w_m$ is:
$$ PMI(w_n;w_m) = \log \dfrac{P(w_n,w_m)}{P(w_n)P(w_m)} $$
where $P(w_n,w_m)$ is estimated counting the number of occurrences of both $w_n$ and $w_m$ in the same visual instance (or subregion of it) and $P(w_n),P(w_m)$ are estimated counting the number of occurrences of $w_n, w_m$ in the whole corpus.
\end{mydef}
Using the point--wise mutual information (PMI) as a measure of associatedness, we can  propose a PMI--based definition of the two other requirements for pairwise synonymy between visual words:  complementary distribution and similar contextual distribution. 
\begin{mydef} [Complementary distribution]
A pair of visual words $w_n, w_m$  have a complementary distribution in the collection when co--occurrence of the two words in the same context is less probable than occurrence of both the words separately. In such a case, the PMI satisfies: $$PMI(w_n;w_m) \leq 0.$$
\end{mydef}

\begin{mydef} [Contextual distribution]
The similarity of the contextual distribution of a pair of words $w_n, w_m$  can be measured by the angle of the vectors $\phi_{n,m} = \angle(\mathbf{d_{n,m}},\mathbf{d_{m,n}})$, where $\mathbf{d_{n,m}}$ is a $(N_W-2)$--dimensional vector where each component is the associatedness  of the word $w_n$ with the word $w_i$, with $i \neq n,m$. Two words $w_n,w_m$ have a similar contextual distribution if 
\begin{eqnarray}
\cos(\phi_{n,m})  \geq  T_{synonymy}
\end{eqnarray}
where 
$$ \cos(\phi_{n,m})  =  \dfrac {\sum_{i \neq n,m} PMI(w_n;w_i) \cdot PMI(w_m;w_i)}{\sqrt{\sum_{i \neq n,m} (PMI(w_n;w_i))^2} \cdot \sqrt{\sum_{i \neq n,m} (PMI(w_m;w_i))^2}}  
$$ and $T_{synonymy} \in [-1,1]$ is the synonymy threshold.  
\end{mydef}

If the conditions of complementary distribution and contextual distribution are met, then the amount of synonymy of two visual words can be quantified.
\begin{mydef} [Synonymy value]
The synonymy value of two words $w_n,w_m$ is the maximum significance value for which both words are significant for the same visual topic. 
\begin{equation}
\sigma_{nm} = \sigma_{mn} = \max_{j} \left\lbrace \min_{n,m}  \left\lbrace t_{n,j},t_{m,j} \right\rbrace \right\rbrace
\end{equation} 
\end{mydef}

The synonymy value enables a transformation of the visual word space considering words with similar meaning but also preserving the choice of one word over the synonyms, since it is informative. 
We propose a synonymy--based transformation of the visual word space, where each transformed word is a linear combination of all its synonyms. 
\begin{mydef}[Synonymy--transformed visual word space]
Let $\mathbf{h}$ be a histogram vector where each component represents the multiplicity of a visual word, and $\mathbf{S}$ a symmetric synonymy transformation matrix:
\begin{eqnarray}
\mathbf{h} & = & (n(w_1),n(w_2),\dots, n(w_{N_W}))^T
\\
\mathbf{S} & = & \left( 
\begin{array}{cccc}
1 & s_{12} & \cdots & s_{1{N_W}} \\ 
s_{21} & 1 & \cdots & s_{{2N_W}} \\ 
\vdots & \vdots & \ddots & \vdots \\ 
s_{{N_W}1} & s_{{N_W}2} & \cdots & 1
\end{array} 
 \right) 
\end{eqnarray}
where $s_{ij}$ measures the synonymy of the visual words $w_i$ and $w_j$.
\begin{equation}
s_{ij} = s_{ji} = \left \lbrace 
  \begin{array}{l}
  	 1 \ \mathrm{if}\ i=j \\
     {\sigma_{ij}} \ \mathrm{if}\ w_i,w_j\  \mathrm{are\ synonyms} \\
     0 \ \mathrm{otherwise} \\
  \end{array}
  \right.
\end{equation}
Then, the vector $\mathbf{h^S} =(n(w^S_1),n(w^S_2),\dots, n(w^S_{N_W}))^T $ is the histogram vector of visual words in the synonymy--transformed space
\begin{eqnarray}
\mathbf{h^S} & = & \mathbf{S}\mathbf{h} \\
n(w^S_i) & = & n(w_i) + \sum_{i\neq j} s_{ij}n(w_j)
\end{eqnarray}
\end{mydef}
Synonymy is a symmetric relation, and so is the transformation matrix. However, by allowing several relations for the same word, two transformed visual words will remain distinguishable as long as they do not share the same synonymy relations at the same levels. This preserves the information contained in the paradigmatic relations.
\subsection{Polysemy transformation}
\label{sec:Polysemy}
As explained in section~\ref{sec:synonymyPolysemyIntroduction}, visual words generated using clustering might produce ambiguous words that are linked to various visual appearances. 
This behaviour is shown by word 5 in Figure~\ref{fig:clusters}. 
Polysemy is the property of a word that has two or more different meanings. 
Polysemic visual words are sources of ambiguity in the description of the visual instances, since they can refer to different visual appearances. 
A visual word $w_n$ is polysemic in wide sense if there are at least two visual topics $z_j,z_k$ to which the visual word belongs. 
Based on the topic--based significance defined in Section~\ref{sec:topicBasedSignificance}, a polysemy threshold for each word can be defined.

\begin{mydef} [Polysemy threshold]
The polysemy threshold of a visual word $w_n$, $T_{polysemy}^n$, is the largest value that satisfies that there are at least two topics for which the word is significant above the threshold:
$$
\left\vert \left\lbrace t_{n,j} \geq T_{polysemy}^n \right\rbrace \right\vert \geq 2; \forall j=1,\dots,N_Z
$$
\end{mydef}

Words with various meanings are ambiguous, which is not a desirable property for a descriptive feature. 

Therefore a transformation of the visual word space should reduce the weight of ambiguous, polysemic words while relatively increasing the weight of specific words with clear meanings. 
\begin{mydef}[Polysemy--transformed visual word space]
Let $\mathbf{h}$ be a histogram vector where each component represents the multiplicity of a visual word, and $\mathbf{P}$ a polysemy transformation matrix:
\begin{eqnarray}
\mathbf{h} & = & (n(w_1),n(w_2),\dots, n(w_{N_W}))^T
\\
\mathbf{P} & = & \left( 
\begin{array}{cccc}
p_1 & 0 & \cdots & 0 \\ 
0 & p_2 & \cdots & 0 \\ 
\vdots & \vdots & \ddots & \vdots \\ 
0 & 0 & \cdots & p_{N_W}
\end{array} 
 \right)
\end{eqnarray}
where $p_{i}$ is the polysemy weight of the visual word $w_i$:
\begin{equation}
p_{i}  =   {1-T_{polysemy}^i}.
\end{equation}
Then, the vector $\mathbf{h^P} =(n(w^P_1),n(w^P_2),\dots, n(w^P_{N_W}))^T $ is the histogram vector of visual words in the polysemy--transformed space.
\begin{eqnarray}
\mathbf{h^P} & = & \mathbf{P}\mathbf{h} \\
n(w^P_i) & = & p_i \cdot n(w_i) 
\end{eqnarray}
\end{mydef}
\subsection{Grammatical similarity}
\label{sec:GrammaticalSimilarity}
The bag of visual words defines a feature space where each dimension is a word, and the components of each feature vector are the occurrences of the word in a visual instance. 
One of the most frequently used approaches for comparing similarity of two vectors in a feature space is computing the distance between two points: if two points are separated by a \emph{small} distance, they are considered similar. Therefore, distances are considered dissimilarity measures. 
The Euclidean distance is the simplest choice but many other distances have been proposed: Manhattan, Bhattacharya, Mahalanobis, etc., each with their own properties and conditions under which they are optimal~\cite{VaL2000}.
Instead of measuring dissimilarity by using a distance, it is possible to use a similarity measure, where \emph{high} values correspond to higher similarities. 
Bullinaria and Levy~\cite{BuL2007} compared various distances and the cosine similarity measure in word similarity. 
The cosine was found the best way of measuring similarity of two bag of words vectors. 
It was also found to be computationally efficient~\cite{Wid2004}. 

Formally, the cosine similarity of two visual instances is the cosine of the angle that their feature vectors define in the bag of visual words space. Let $\mathbf{h_{i}}$ and $\mathbf{h_{j}}$ be two histogram vectors of the bag of visual words of the images $I_i$ and $I_j$, each with $N_W$ elements.
\begin{eqnarray}
\mathbf{h_{i}} & = & (n_i(w_1),n_i(w_2),\dots, n_i(w_{N_W}))
\\
\mathbf{h_{j}} & = & (n_j(w_1),n_j(w_2),\dots, n_j(w_{N_W}))
\end{eqnarray}
The cosine similarity between the visual instances $I_i$ and $I_j$ is: 
\begin{eqnarray}
sim_{\cos}(I_i,I_j) &  = & \dfrac{\mathbf{h_{i}} \cdot \mathbf{h_{j}}}{\left\Vert \mathbf{h_{i}} \right \Vert \left\Vert \mathbf{h_{j}} \right \Vert} \\
&  = & \dfrac{\sum_{k=1}^{N_W} n_i(w_k) \cdot n_j(w_k)}{\sqrt{\sum_{k=1}^{N_W} (n_i(w_k))^2 \cdot \sum_{k=1}^{N_W} (n_j(w_k))^2 }}
\end{eqnarray}
The use of the cosine similarity has two important advantages:
\begin{itemize}
\item It is a bounded measure of similarity, which takes the value 1 for histograms pointing in the same direction and 0 for orthogonal directions. The $-1$ value, meaning vectors with opposite directions does not take place in the pure bag of visual words model, where vector components are positive by definition. 
\item It is not biased by the absolute number of visual words in a visual instance, since only the relative direction is considered. 
\end{itemize}
Based on the use of the cosine similarity and the various transformations to the feature space, we can define the grammatical similarity of two visual instances.
\begin{mydef} (Grammatical similarity)
The grammatical similarity between two visual instances $I_i$ and $I_j$, represented by the histogram vectors $h_i$ and $h_j$ is:
\begin{equation}
sim_{gram}(I_i,I_j)=\dfrac{\mathbf{(M \cdot P \cdot S \cdot h_i)^T \cdot (M \cdot P \cdot S \cdot h_j)}}{\left\Vert \mathbf{(M \cdot P \cdot S \cdot h_i})\right\Vert \cdot \left\Vert \mathbf{(M \cdot P \cdot S \cdot h_j})\right\Vert}
\end{equation}
\end{mydef}
This similarity measure has the advantages of the cosine measure, but also considers the relative properties of visual words based on their behaviour in the training data. 
\section{Evaluation}
\label{sec:Evaluation}
In this section we describe the results of experimental evaluation of the visual grammar approach. 
\subsection{Pascal VOC challenge}
\label{sec:PASCAL}
 According to the challenge website\footnote{\url{http://host.robots.ox.ac.uk/pascal/VOC/voc2007/}}, the goal of the Visual Object Classes (VOC) challenge is to recognize objects from 20 visual object classes in realistic scenes (i.e. not pre--segmented objects). It is fundamentally a supervised learning learning problem in that a training set of labelled images is provided.
 
The results on the Pascal data set were obtained using the feature encoding evaluation framework provided by Chatfield et al.~\cite{CLV2011}, since it allows the comparison with other methods. Table~\ref{tab:VOC2007} shows the results obtained using the visual grammar transformation without adding the cosine similarity metric (which is not supported by the toolkit). 

\begin{table}
    
    \begin{center}
    \begin{tabular}{|lll|}
    \hline
    Method & Descriptor length & MAP(\%) \\
    \hline
    VG (0.9)                & 4824               & 46.93   \\
    VG (0.8)                      & 16504              & 52.57   \\
    VG (0.7)                      & 21703              & 53.49   \\
    VG (0.6)                      & 24029              & 53.61   \\
    VG (0.5)                      & 24804              & 53.46   \\
    VG (0.4)                      & 24964              & 53.60   \\
    VG (0.3)                      & 24983              & 53.83   \\
    VG (0.2)                      & 24990              & 53.97   \\
    VG (0.1)                      & 24995              & 53.85   \\ 
    \hline
    BoVW baseline            & 25000              & 53.85   \\
    Fisher* & 41000 & 61.69 \\  
    Super Vector* & 132000 & 58.16 \\ \hline
    \end{tabular}
    \end{center}
    \label{tab:VOC2007}
     \caption{Mean average precision on the VOC2007 dataset, compared to other methods. The results from the Fisher and Super Vector encodings are those reported by Chatfield et al.~\cite{CLV2011}} \caption{Mean average precision on the VOC2007 dataset, compared to other methods. The results from the Fisher and Super Vector encodings are those reported by Chatfield et al.~\cite{CLV2011}, whereas the BoVW was computed using the toolkit without enabling fine tuning features}
\end{table}
\subsection{ImageCLEF modality classification}
\label{sec:ImageCLEF}
In order to extend the evaluation including the cosine similarity, the ImageCLEF modality classification dataset was used. 

Image modality is one of the characteristics of medical image retrieval that practitioners would like to see included in existing systems~\cite{MHD2012}.
Medical image search engines such as GoldMiner\footnote{\url{http://goldminer.arrs.org/}} and Yottalook\footnote{\url{http://www.yottalook.com/}} contain modality filters to improve retrieval results. 
Whereas DICOM headers often contain meta data that can be used to filter modalities, this information is lost when exporting images for publication in journals or conferences where images are stored as JPG, GIF or PNG files.
In this case visual appearance is key to identify modalities or the caption text can be analyzed for respective keywords to identify modalities. 
The ImageCLEFmed evaluation campaign contains a modality classification task that is regarded as an essential part for image retrieval systems.
In 2013, the modality classification training set contained 2,896 images from the medical literature organized in a hierarchy of 31 image types~\cite{MKD2012}. 

Images were described with a BoVW based on SIFT~\cite{Low2004} descriptors.
This representation has been commonly used for image retrieval because it can be computed efficiently~\cite{MGE2012,YaN2010,KeS2004}.
The SIFT descriptor is invariant to translations, rotations and scaling transformations and robust to moderate perspective transformations and illumination variations.
SIFT encodes the salient aspects of the grey--level images gradient in a local neighborhood around each interest point.
 
\begin{figure}[bt]
\centering
\includegraphics[width=10cm]{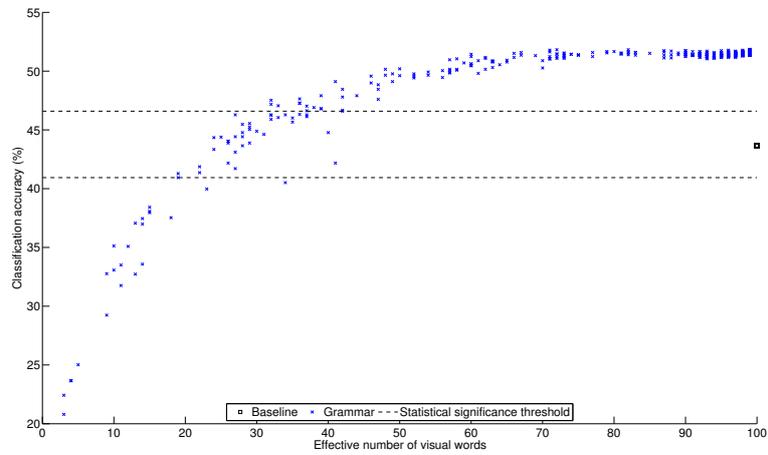}
\caption{Classification accuracy versus effective vocabulary size, compared to the baseline approach using a 1--nearest neighbor classifier on an initial vocabulary of 100 words. }
\label{fig:sift100accuracy}
\end{figure}

\begin{figure}[bt]
\centering
\includegraphics[width=10cm]{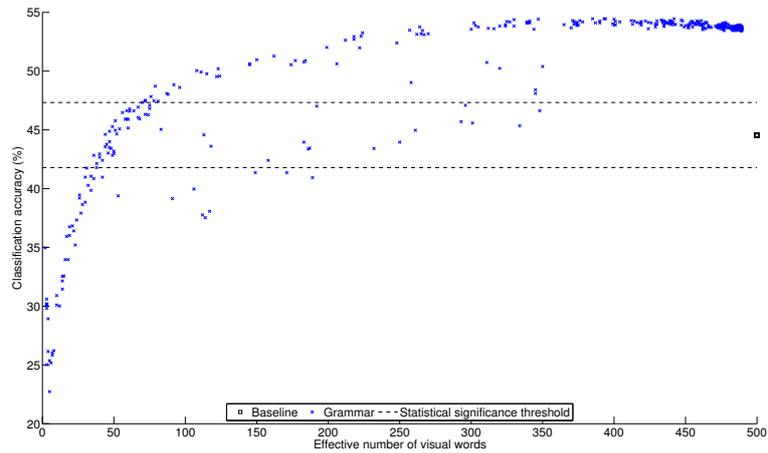}
\caption{Classification accuracy versus effective vocabulary size, compared to the baseline approach using a 1--nearest neighbor classifier on an initial vocabulary of 500 words. }
\label{fig:sift500accuracy}
\end{figure}

Evaluation with separate training and test sets was performed using all combinations of the following parameters:
\begin{enumerate}
 \item Two SIFT--based visual vocabularies with 100 and 500 visual words.
 \item A varying number of visual topics from 25 to 350 in steps of 25. 
 \item A varying meaningfulness threshold from 50\% to 100\%.
 \item A K--NN classifier with $K$ values 1, 5 and 10.
\end{enumerate} 
%
Figures~\ref{fig:sift100accuracy} and~\ref{fig:sift500accuracy} show the results obtained with all configurations for each vocabulary using a 1--NN classifier.

\section{Discussion}
\label{sec:Discussion}
The impact of the visual grammar approach on classification and retrieval tasks lie on two areas: first, an increase in the classification accuracy and second, a reduction of the descriptor size. 

According to the relative accuracy with respect to the baseline, three effects can be discussed in terms of the descriptor size. 
\begin{itemize}
\item Stable, statistically significant improved accuracy ($p<0.05$) was obtained for small reductions of the vocabulary size when using the cosine similarity together with the visual grammar transformation. 
\item Similar accuracies to the baseline approach were obtained while maintaining a moderate to large reduction of the vocabulary size regardless of the similarity metric used.  
\item Extreme reductions beyond 10-20\% of the initial vocabulary size significantly reduced accuracy as well as vocabulary size. 
\end{itemize}
\section{Conclusions}
\label{sec:Conclusions}
In this paper we present a conceptual analysis of three major properties of language grammar and how they can be adapted to the computer vision and image understanding domain based on the bag of visual words paradigm. 
Meaningfulness of visual words is quantified for dimensionality reduction or feature weighting.
Synonymy is modelled according to a set of criteria that enable defining relations between pairs of visual words and quantifying them.
Polysemy of visual words is also quantified and identified as a source of ambiguity.

These properties are able to define three transformations of the standard bag of visual words space, and a similarity measure based on the cosine is proposed, incorporating all the transformations. 

Evaluation of the visual grammar shows that a positive impact on classification accuracy and/or descriptor size when the proposed techniques are applied. 

The visual grammar transformation only outperforms recent methods such as the Fisher vectors and super vector encoding in descriptor size but cannot provide better accuracy if it is not used in combination with the cosine similarity. It also provides a framework that can identify relations between features from different types as well as reducing dimensionality of the descriptors.

Future work includes combining various types of features to identify relationships among vocabularies of different nature and exploring in depth the interactions among the various proposed transformations. 
\section*{Acknowledgements}
This work was partially supported by the Swiss National Science Foundation (FNS) in the MANY2 project (205320--141300), the EU $7^{th}$ Framework Program under grant agreements  257528 (KHRESMOI) and 258191 (PROMISE).

\end{document}